\documentclass{article}
\usepackage{ijcai22}

\usepackage{times}
\usepackage{soul}
\usepackage{url}
\usepackage[hidelinks]{hyperref}
\usepackage[utf8]{inputenc}
\usepackage[small]{caption}
\usepackage{graphicx}
\usepackage{amsmath}
\usepackage{amsthm}
\usepackage{booktabs}
\usepackage{algorithm}
\usepackage{algorithmic}
\usepackage{lineno}
\usepackage{bm}
\usepackage{bbm}
\usepackage{float}
\usepackage{multirow}
\usepackage{makecell}
\usepackage{tabularx}
\usepackage{amssymb}
\usepackage{stmaryrd}
\usepackage{lipsum}
\usepackage{subfigure}
\usepackage{bbold}

\title{Bootstrapping Informative Graph Augmentation via A Meta Learning Approach}

\author{
Hang Gao$^{1,2}$\footnotemark[1]
\and
Jiangmeng Li$^{1,2}$\footnote{Contributed equally to this work, in no particular order.}\and
Wenwen Qiang$^{1,2}$\footnote{Corresponding author}\and
Lingyu Si$^{1,2}$\and
Fuchun Sun$^{3}$\and\\
Changwen Zheng$^{2}$
\affiliations
$^1$University of Chinese Academy of Sciences\\
$^2$Institute of Software Chinese Academy of Sciences\\
$^3$Tsinghua University\\
\emails
\{gaohang, jiangmeng2019, wenwen2018, lingyu, changwen\}@iscas.ac.cn,
fcsun@tsinghua.edu.cn
}

\begin{document}

\maketitle

\begin{abstract}
Recent works explore learning graph representations in a self-supervised manner. In graph contrastive learning, benchmark methods apply various graph augmentation approaches. However, most of the augmentation methods are non-learnable, which causes the issue of generating unbeneficial augmented graphs. Such augmentation may degenerate the representation ability of graph contrastive learning methods. Therefore, we motivate our method to generate augmented graph with a learnable graph augmenter, called MEta Graph Augmentation (MEGA). We then clarify that a "good" graph augmentation must have uniformity at the instance-level and informativeness at the feature-level. To this end, we propose a novel approach to learning a graph augmenter that can generate an augmentation with uniformity and informativeness. The objective of the graph augmenter is to promote our feature extraction network to learn a more discriminative feature representation, which motivates us to propose a meta-learning paradigm. Empirically, the experiments across multiple benchmark datasets demonstrate that MEGA outperforms the state-of-the-art methods in graph self-supervised learning tasks. Further experimental studies prove the effectiveness of different terms of MEGA. Our codes are available at https://github.com/hang53/MEGA. 
\end{abstract}

\section{Introduction}
Recently, there has been a surge of interest in learning a graph representation via self-supervised Graph Neural Network (GNN) approaches. GNNs, inheriting the powerful representation capability of neural networks, emerged as benchmark approaches over many graph representation learning tasks. Early works mostly require task-dependent labels to learn a graph representation. However, annotating graphs is a rather challenging task compared to labeling common modalities of data, especially in specialized domains. Therefore, recent research efforts are dedicated to developing self-supervised graph representation learning methods, which can eliminate the dependency of the labels \cite{hu2020strategies}.

Graph contrastive learning (GCL), one of the most popular self-supervised methods in graph representation learning, is proposed based on GNNs and contrastive learning. Under the learning paradigm of contrastive learning, GCL generates augmented graphs by adopting graph augmentation \cite{hassani2020contrastive}. After graph encoding, the augmented and original features of the same graph are treated as positives, and the features of different graphs are treated as negatives. The object of GCL is to learn a good graph representation by pulling the positives close and pushing the negatives apart. However, most of the graph augmentation approaches are non-learnable, which causes two issues: 1) the augmentation is excessively weak, e.g., the augmented graph is indistinguishable from the original graph, and the contrastive learning model can hardly mine consistent knowledge from them; 2) the augmentation introduces overmuch noise, and the augmented graph is even more similar to other graphs. The mentioned issues weaken GCL's ability to learn a discriminative representation. Therefore, we motivate our method to directly \textit{learn} a graph augmentation, which can assist GCL in generating a good graph representation.

We aim to learn a "good" graph representation that can have impressive performance on downstream tasks, but what is an exact "good" representation? From \cite{2020Bootstrap} \cite{2020WhiteningErmolov}, we notice that, at the instance-level, a good representation naturally has \textit{uniformity}, e.g., features of different samples are scattered throughout the hidden space instead of collapsing to a point. However, such constraint does not consider the representation's collapse at the feature-level. For instance, the learned representation has 256 dimensions, but most of them have few differences, which implies that much information learned by the representation is redundant \cite{2021Barlow}. Such redundant information may lead to limited informativeness of the representation and degenerate the representation to model truly discriminative information. Therefore, we motivate our method to learn a "good" representation with uniformity at the instance-level and informativeness at the feature-level.

\begin{figure}
	\centering
	\includegraphics[width=0.45\textwidth]{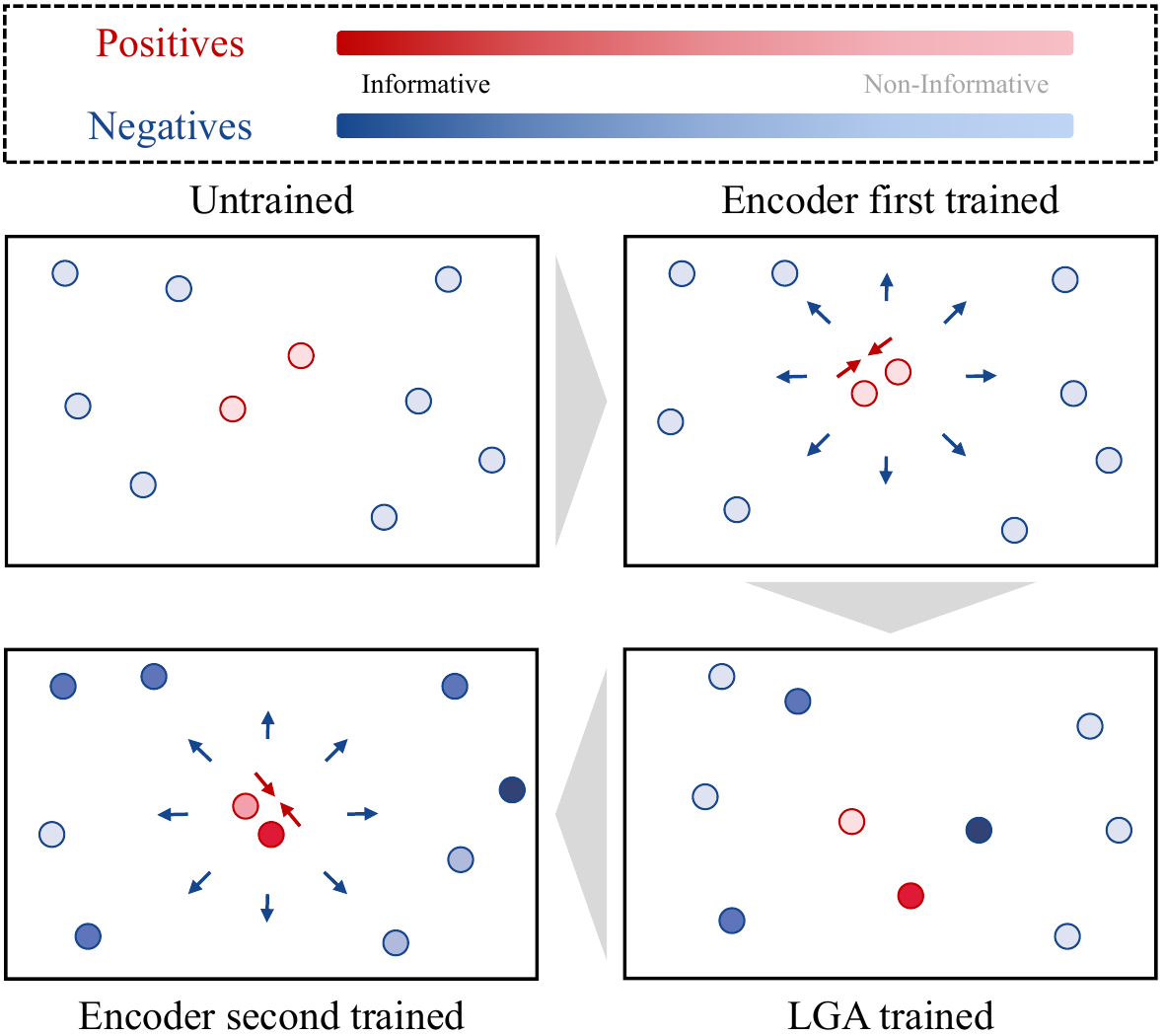}
	\caption{An illustration example of the training process of MEGA. The figure shows the features in hidden space during training. Red points depicts the positive features and blue points depicts the negative features at the instance-level. The gradation of color denotes the informativeness at the feature-level.}
	\label{fig:example}
\end{figure}

To this end, we propose \textbf{ME}ta \textbf{G}raph \textbf{A}ugmentation (\textbf{MEGA}) to guide the encoder to learn a discriminative and informative graph representation. For training the encoder, we follow the common setting of contrastive learning \cite{chen2020simple}. The well self-supervised contrastive learning approach leads the features to be scattered in the hidden space. However, in practice, the sufficient self-supervision of the contrastive learning approach demands hard (informative at the instance-level) features of positive and negative samples, e.g., positive features that are relatively far apart and negative features that are relatively close in the hidden space. For instance, \cite{chen2020simple} leverages large batch and memory bank to include more hard features, \cite{chuang2020debiased} explore to emphasize hard features in training. At the instance-level, our motivation is to straightforwardly generate hard graph features by a learnable graph augmenter (LGA), which uses the encoder's performance in one iteration to generate hard features for the next iteration by updating the graph augmentation. Note that the objective of LGA is \textit{not} to promote convergence of contrastive loss. On the contrary, we expect LGA to generate an augmented graph that can increase the difficulty of the self-supervision problem (i.e., contrasting). Contrastive learning aims to put the original and augmented features of a graph together and push the features of different graphs away, and LGA aims to degenerate such a process. Therefore, the LGA augmented graph feature must be hard for contrastive learning. To ensure the informativeness of the learned representation at the feature-level, we propose to train LGA to augment the graph so that it can improve the encoder to generate a representation with informativeness at the feature-level. As shown in Figure \ref{fig:example}, LGA is like a teacher that shows different hard and informative examples (augmented graphs) to the encoder, and the contrastive loss leads the encoder to learn discriminative knowledge from them.

The reason why we take a meta-learning approach to update LGA is as follows: the learning paradigm of meta-learning ensures that the optimization objective of LGA is improving the \textit{encoder} to learn representations with uniformity at the instance-level and informativeness at the feature-level from graphs. However, a regular learning paradigm, e.g., directly optimizing LGA by the loss of measuring the uniformity and informativeness of features in hidden space, can only ensure that the features learned from the augmented graph are modified. However, the features learned from the original graph could be collapsed or non-informative. Concretely, the meta-learning paradigm ensures that the encoder learns the knowledge to generate good representations with uniformity at the instance-level and informativeness at the feature-level.

\textbf{Contributions.} The takeaways of this paper are as follows:

\begin{itemize}
	\item We propose a learnable approach to generate informative graph augmentation, called meta graph augmentation, which boosts the performance of graph contrastive learning.
	\item We propose an auxiliary meta-learning approach to train the learnable graph augmenter, which guides the encoder to learn a representation with uniformity at the instance-level and informativeness at the feature-level.
	\item We conduct experiments to compare our method with state-of-the-art graph self-supervised learning approaches on benchmark datasets, and the results prove the superiority of our method.
\end{itemize}

\begin{figure*}[t]
	\centering
	\includegraphics[width=1.0\textwidth]{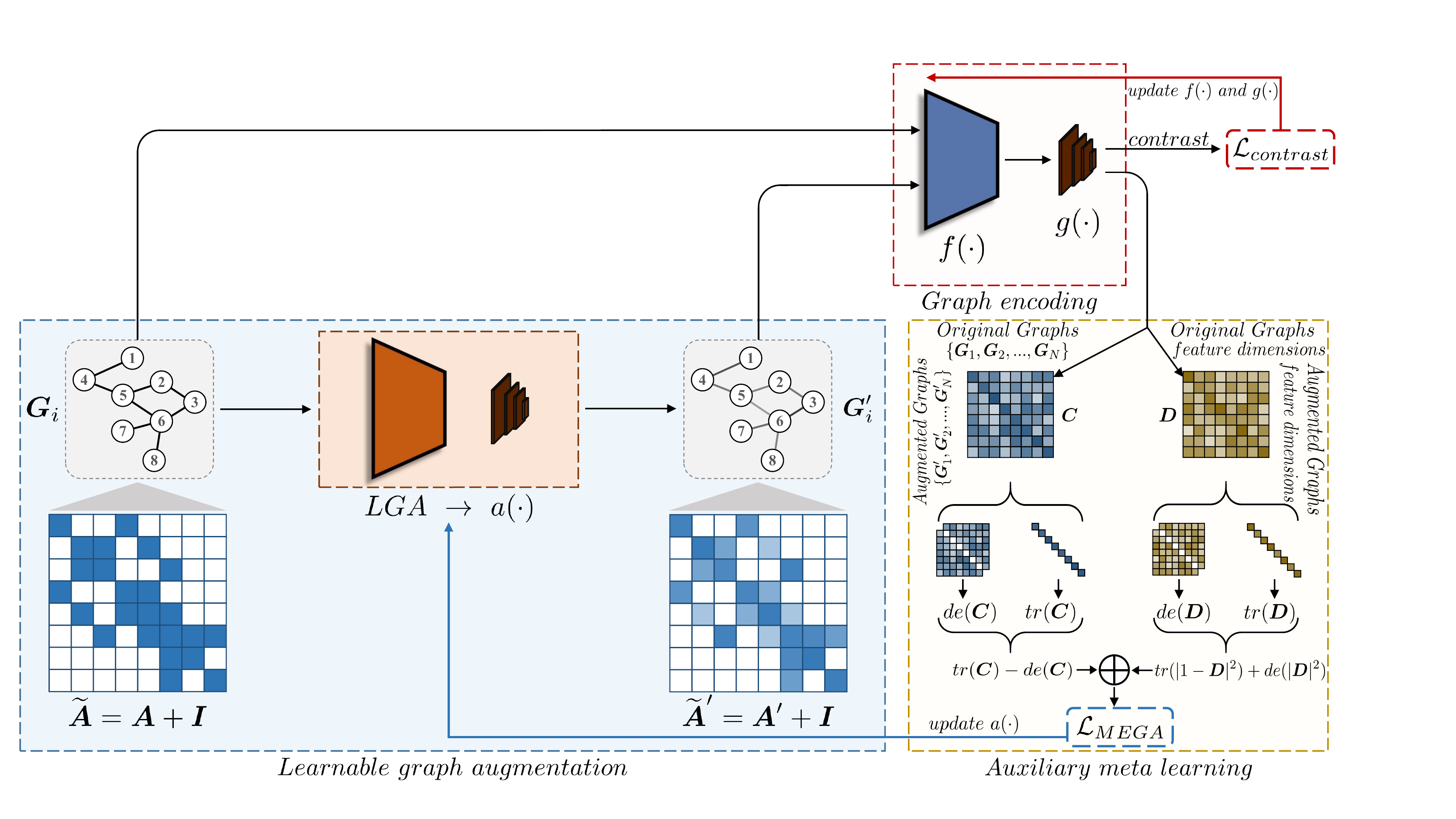}
	\caption{MEGA's architecture. MEGA uses LGA to generate augmented graph, which and the original graph are encoded together. In one iteration, the encoder and projection head are trained by back-propagating $\mathcal{L}_{contrast}$, and in the next iteration, the LGA is trained by performing the second-derivative technique on $\mathcal{L}_{MEGA}$. The encoder is trained until convergence.}\label{fig:algoframe}
\end{figure*}

\section{Related Works}
In this section, we review some representative works on graph neural networks, graph contrastive learning, and meta-learning, which are related to this article.

\paragraph{Graph Neural Networks (GNN).} GNN can learn the low-dimensional representations of graphs by aggregating neighborhood information. These representations can then be applied to various kinds of downstream tasks.
Like other neural network structures, GNNs developed many variants. Graph Convolution Networks (GCNs) \cite{kipf2016semi}, as an extension of the convolutional neural network on graph-structured data, use convolution operation to transform and aggregate features from a node's graph neighborhood. \cite{xu2018powerful} shows that GNNs are at most as powerful as the Weisfeiler-Lehman test in distinguishing graph structures. Based on this idea, \cite{xu2018powerful} proposed Graph Isomorphism Networks (GINs). Graph Attention Networks (GATs) \cite{velivckovic2017graph} introduces attention mechanisms into graph learning.

\paragraph{Contrastive learning.} Contrastive Learning is a kind of self-supervised learning approach that measures the loss in latent space by contrasting features in hidden space. CMC \cite{tian2020contrastive} uses multi-view data to acquire features for contrasting. In computer vision, many works based on contrastive learning have achieved outstanding results in different kinds of tasks \cite{chen2020simple} \cite{2021Barlow}. As in graph learning, contrastive learning also has many applications. For instance, DGI \cite{velivckovic2018deep} learns node representations through contrasting node and graph embeddings. \cite{hassani2020contrastive} learns node-level and graph-level representations by contrasting the different structures of a graph.

\paragraph{Meta learning.} The objective of meta-learning is to learn the \textit{learning algorithm} automatically. Early works \cite{2014Schmidhuber} aim to guide the model (e.g., neural network) to learn prior knowledge about \textit{how to learn new knowledge}, so that the model can efficiently study new information. Recently, researchers explored using meta-learning to find optimal hyper-parameters and appropriately initialize a neural network for few-shot learning \cite{2017ModelFinn}. 

\section{Methods}
In this section, we introduce the proposed \textbf{ME}ta \textbf{G}raph \textbf{A}ugmentation (\textbf{MEGA}). The architecture of MEGA is depicted in Figure \ref{fig:algoframe}. MEGA proposes to learn informative graph augmentation by a meta-learning approach. Guided by the augmented graph, the GNN encoder can mine hidden discriminative knowledge from the original graph.

\subsection{Preliminary}
We recap necessary preliminary concepts and notations for further exposition. In this paper, we
consider attributed graphs $\boldsymbol{G} = (\boldsymbol{V}, \boldsymbol{E})$ where $\boldsymbol{V}$ is a node set and $\boldsymbol{E}$ is the corresponding edge set. For $\boldsymbol{G}$, $\{\boldsymbol{X_v} \in \mathbb{R}^{\boldsymbol{V}} | \boldsymbol{v} \in \boldsymbol{V}\}$ denotes the node attributes.

\paragraph{Learning graph representations.} Given a graph dataset $\mathcal{G}$ including $\boldsymbol{G}_i$, where $i \in \llbracket{1, N} \rrbracket$. Our objective is to learn an encoder $f(\cdot): \mathcal{G} \to \mathbb{R}^{\boldsymbol{D}}$, where $f(\boldsymbol{G}_i)$ is a representation that contains discriminative information of $\boldsymbol{G}_i$ and can be further used in downstream
task. We assume $\boldsymbol{G}_i$ as a random variable that is sampled \textit{i.i.d} from distribution $\mathcal{P}\left(\mathcal{G} \right)$ defined over $\mathcal{G}$.
To learn such discriminative representation $f(\boldsymbol{G}_i)$, we adopt GNN as the encoder and then perform self-supervised contrastive learning in hidden space.

\paragraph{Graph Neural Networks.} In this paper, we focus on using GNN, message passing GNN in particular, as the encoder $f(\cdot)$. For graph $\boldsymbol{G}_i = (\boldsymbol{V}_i, \boldsymbol{E}_i)$, we denote $\boldsymbol{H}_{\boldsymbol{v}}$ as the representation vector, for each node $\boldsymbol{v}\in \boldsymbol{V}_i$. The $k$-th layer GNN can be formulated as:
\begin{align}
	{\boldsymbol{H}^{(k+1)}_{\boldsymbol{v}}} = combine^{(k)} \Big(\boldsymbol{H}^{k}_{\boldsymbol{v}},\nonumber aggregate^{(k)} (\boldsymbol{H}^{k}_{\boldsymbol{u}}\nonumber\\
	,\forall \boldsymbol{u} \in \mathcal{N}(\boldsymbol{v})  ) \Big),
	\label{eq:gnn}
\end{align}
where $\mathcal{N}(\boldsymbol{v})$ denotes the neighbors of node $\boldsymbol{v}$, $\boldsymbol{H}^{(k)}$ is the representation vector of the node $\boldsymbol{v}$ at layer $k$, and when $k = 0$, $\boldsymbol{H}^{(0)}$ is initialized with the input node features, which is extracted from $\boldsymbol{X}$. $combine$ and $aggregate$ are functions with learnable parameters. After $K$ rounds of massage passing, a readout function will pool the node representations to obtain the graph representation $\boldsymbol{h}_i$ for $\boldsymbol{G}_i$:
\begin{align}
	{\boldsymbol{h}_{i}} = readout \Big(\boldsymbol{H}_{\boldsymbol{v}}, \boldsymbol{v} \in \boldsymbol{V}_i\Big).
	\label{eq:gnn}
\end{align}

\paragraph{Contrastive learning.} We follow the preliminaries of contrastive learning \cite{tian2020contrastive}: learning an embedding that maximizes agreement between the original and augmented features of the same sample, namely \textit{positives}, and separates the features of different samples, namely \textit{negatives}, in latent space. We denote $\boldsymbol{G}^\prime_i$ is the augmented graph of $\boldsymbol{G}_i$.


To impose contrastive learning, we feed the inputs $\boldsymbol{G}_i$ and $\boldsymbol{G}^\prime_i$ into the encoder $f(\cdot)$ to learn representations $\boldsymbol{h}_i$ and $\boldsymbol{h}_i^\prime$, and the representations are further mapped into features $\boldsymbol{z}_i$ and $\boldsymbol{z}_i^\prime$ by a projection head $g(\cdot)$ \cite{chen2020simple}. The encoder $f(\cdot)$ and projection head $g(\cdot)$ are trained by contrasting the features, and the loss \cite{2018RepresentationOord} is formulated as follows:
\begin{equation}
	{\mathcal{L}_{contrast}} = -\log \frac{\exp\left(\frac{<\boldsymbol{z}^+>}{\tau}\right)}{\exp\left(\frac{<\boldsymbol{z}^+>}{\tau}\right) + \sum {\exp\left(\frac{<\boldsymbol{z}^{-}>}{\tau}\right)}}
	\label{eq:cl}
\end{equation}
where $\boldsymbol{z}^+$ denotes the pair of $\{\boldsymbol{z}_i^\prime, \boldsymbol{z}_i\}$, $\boldsymbol{z}^-$ is a set of pairs, i.e., $\Big\{\{\boldsymbol{z}_j^\prime, \boldsymbol{z}_i\}, \{\boldsymbol{z}_i^\prime, \boldsymbol{z}_j\} \Big| j \in \llbracket{1, N} \rrbracket, j \neq i \Big\}$, $<\cdot>$ denotes a discriminating function to measure the similarity of features, and $\tau$ is the temperature factor \cite{chen2020simple}. Note that, after training is completed, the projection head $g(\cdot)$ is discarded, and the representations are directly used for downstream tasks.

\subsection{Meta Graph Augmentation}
Different from benchmark methods that randomly dropping or adding edges to augment a graph \cite{hassani2020contrastive} \cite{wan2020contrastive}, we motivate our method to impose graph augmentation in an learnable manner \cite{suresh2021adversarial}. We rethink the learning paradigm of contrastive learning and find that such an approach relies heavily on hard and informative features in training. Therefore, to boost the performance on downstream tasks of the learned representations, we propose to use a trick of meta-learning to generate informative graph augmentation, which is to guide the encoder to mine discriminative knowledge from graphs.

\paragraph{Learnable graph augmentation.} As the architecture shown in Figure \ref{fig:algoframe}, we propose a learnable approach to augment graph. In detail, suppose $\boldsymbol{A}$ is the adjacency matrix of $\boldsymbol{G}_i$ where the initial weights of the connected nodes are set to $1$ and others are valued by $0$. $\widetilde{\boldsymbol{A}} = \boldsymbol{A} + \boldsymbol{I}$, where $\boldsymbol{I}$ denotes the self-connection of each node $\boldsymbol{e} \in \boldsymbol{E}$. We use a neural network $a(\cdot)$, as the LGA (see Appendix for the implementation), to generate the augmented graph $\boldsymbol{G}_i^\prime$ from the original graph $\boldsymbol{G}_i$, where $\widetilde{\boldsymbol{A}}_i^\prime$ is the adjacency matrix with self-connections $\boldsymbol{I}$ of a graph $\boldsymbol{G}_i^\prime$. $\boldsymbol{G}_i$ and $\boldsymbol{G}_i^\prime$ are then encoded into features $\boldsymbol{z}_i$ and $\boldsymbol{z}_i^\prime$ by the encoder $f(\cdot)$ and projection head $g{(\cdot)}$.

\begin{table*}\normalsize
	
	
	\setlength{\tabcolsep}{5pt}
	\begin{center}
		\begin{tabular}{l|cccccccc}
			\hline\rule{0pt}{10pt}
			{Method} & PROTEINS & MUTAG & DD & COLLAB 
			& RDT-M5K & IMDB-B & IMDB-M  \\
			\hline
			\text{GIN RIU}& 69.03$\pm0.33$ & 87.61$\pm0.39$ & 74.22$\pm0.30$ & 63.08$\pm0.10$ & 27.52$\pm0.61$ & 51.86$\pm0.33$ & 32.81$\pm0.57$\\
			\text{InfoGraph} &  72.57$\pm0.65$ &  87.71$\pm1.77$ & 75.23$\pm0.39$ &  70.35$\pm 0.64$ & 51.11$\pm0.55$ &  71.11$\pm0.88$ &  48.66$\pm0.67$  \\
			\text{GraphCL} &  72.86$\pm1.01$ &  88.29$\pm1.31$ &  74.70$\pm0.70$ &  71.26$\pm 0.55$ &  53.05$\pm0.40$ &  70.80$\pm0.77$ &  48.49$\pm0.63$  \\
			\text{AD-GCL} &  73.46$\pm0.67$ &  89.22$\pm1.38$ &  74.48$\pm0.62$ &  72.90$\pm0.83$ & 53.15$\pm0.78$ &  71.12$\pm0.98$ &  48.56$\pm0.59$   \\
			\hline
			\textbf{MEGA-IL}  &  73.89$\pm0.62$ & 90.34$\pm1.20$ &  \bf75.78$\bf\pm0.63$ &  73.54$\pm0.82$ & 53.16$\pm0.65$ & 71.08$\pm0.73$ &  49.09$\pm0.79$   \\
			\textbf{MEGA}  &  \bf74.20$\bf\pm0.73$ &  \bf91.10$\bf\pm1.34$ &  75.56$\pm0.63$ &  \bf73.96$\bf\pm0.73$ & \bf54.32$\bf\pm0.79$ & \bf71.95$\bf\pm0.98$ &  \bf49.52$\bf\pm0.62$   \\
			\hline
		\end{tabular}
	\end{center}
	\caption{Performance of classification accuracy on datasets from TU Dataset (Averaged accuracy $\pm$ std. over 10 runs). We highlight the best records in bold.}
	\label{tab:test_tu}
\end{table*}

\begin{table*}\normalsize
	
	
	\setlength{\tabcolsep}{7pt}
	\begin{center}
		\begin{tabular}{l|cc|ccc}
			\hline\rule{0pt}{10pt}

			\multirow{2}*{Method}  & molesol & mollipo  &  molbbbp   & moltox21 &  molsider \\
			\cline{2-6}\rule{0pt}{10pt}
			& \multicolumn{2}{c|}{Regression tasks (RMSE $\downarrow$)} & \multicolumn{3}{c}{Classification tasks (ROC-AUC \% $\uparrow$)} \\ 	
			\hline
			\text{GIN RIU}  &1.706$\pm0.180$ & 1.075$\pm0.022$ & 64.48$\pm2.46$  & 71.53$\pm0.74$ & 62.29$\pm1.12$ \\
			
			\text{InfoGraph} &  1.344$\pm0.178$ & 1.005$\pm0.023$   &  66.33$\pm2.79$ &  69.74$\pm0.57$  & 60.54$\pm0.90$  \\
			\text{GraphCL} &   1.272$\pm0.089$ & 0.910$\pm0.016$  & 68.22$\pm1.89$ &  72.40$\pm1.01$ & 61.76$\pm1.11$  \\
			\text{AD-GCL} & 1.270$\pm0.092$ & 0.926$\pm0.037$ & 68.26$\pm1.32$ &  71.08$\pm0.93$ & 61.83$\pm1.14$  \\
			\hline
			\textbf{MEGA-IL}  & 1.153$\pm0.103$ &  0.852$\pm0.022$  &  68.34$\pm1.38$  &  72.08$\pm0.82$ &  \bf63.37$\bf\pm0.87$   \\
			
			\textbf{MEGA}  & \bf1.121$\bf\pm0.092$ &  \bf0.831$\bf\pm0.018$  &  \bf69.71$\bf\pm1.56$ &  \bf72.45$\bf\pm0.67$ &  62.92$\pm0.76$   \\
			\hline
		\end{tabular}
	\end{center}
	\caption{Performance of chemical molecules property prediction in OGB datasets. There are two kinds of tasks, regression tasks and classification tasks. We highlight the best records in bold.  }
	\label{tab:test_ogbg}
\end{table*}

\paragraph{Auxiliary meta-learning.} In contrastive learning, we need hard and informative features to learn discriminative representations. To this end, we build an LGA and train it by a meta-learning approach. In training, we first fix LGA $a_{\sigma}(\cdot)$ and train the encoder $f_{\phi}(\cdot)$ and the projection head $g_{\varphi}(\cdot)$ by back-propagating $\mathcal{L}_{contrast}$, where $\sigma$, $\phi$, and $\varphi$ denote the network parameters of $a(\cdot)$, $f(\cdot)$, and $g(\cdot)$, respectively. Then, $f_{\phi}(\cdot)$ and $g_{\varphi}(\cdot)$ are fixed, and $a_{\sigma}(\cdot)$ is trained by computing its gradients with respect to the performance of $f_{\phi}(\cdot)$ and $g_{\varphi}(\cdot)$, and the meta-learning updating objective is as follows:
\begin{equation}
	\mathop{\arg\min}_{\sigma} \bigg( \mathcal{L}_{MEGA}\Big(g_{\mathring{\varphi}}\big(f_{\mathring{\phi}}( \boldsymbol{G} )\big), g_{\mathring{\varphi}}\big(f_{\mathring{\phi}}(a_\sigma( \boldsymbol{G} ))\big) \Big) \bigg)
	\label{eq:metasigmaj}
\end{equation}
where $g_{\mathring{\varphi}}\big(f_{\mathring{\phi}}( \boldsymbol{G} )\big)$ denotes a set of the features extracted from original graphs, $g_{\mathring{\varphi}}\big(f_{\mathring{\phi}}(a_\sigma( \boldsymbol{G} ))\big)$ denotes a set that includes the features of augmented graphs, and $a_\sigma( \boldsymbol{G} )$ denotes $\boldsymbol{G}^\prime$. $\mathring{\phi}$ and $\mathring{\varphi}$ represent the corresponding parameters of the encoder and projection head, which are updated with one gradient back-propagation using the contrastive loss defined in Equation \ref{eq:cl}:
\begin{equation}
	\begin{aligned}
		&\mathring{\phi} = \phi - \ell\nabla_{\phi}\bigg( \mathcal{L}_{contrast}\Big(g_\varphi\big(f_\phi( \boldsymbol{G} )\big), g_{\varphi}\big(f_{\phi}( \boldsymbol{G}^\prime )\big) \Big) \bigg)\\
		&\mathring{\varphi} = \varphi - \ell\nabla_{\varphi}\bigg( \mathcal{L}_{contrast}\Big(g_\varphi\big(f_\phi( \boldsymbol{G} )\big), g_{\varphi}\big(f_{\phi}( \boldsymbol{G}^\prime )\big) \Big) \bigg)
	\end{aligned}
	\label{eq:fastweight}
\end{equation}
where $\ell$ is the learning rate shared between $\phi$ and $\varphi$. The idea behind the meta updating objective is that we perform the second-derivative trick \cite{2019SelfLiu} to train $a_{\sigma}(\cdot)$. Specifically, a derivative over the derivative (i.e., a Hessian matrix) of $\{\phi, \varphi\}$ is used to update $\sigma$. We compute the derivative with respect to $\sigma$ by using a retained computational graph of $\{\phi, \varphi\}$. Then, $\sigma$ is updated by
\begin{equation}
	\begin{aligned}
		\sigma = \sigma - \ell^\prime\nabla_{\sigma}\bigg( \mathcal{L}_{MEGA}\Big(g_{\mathring{\varphi}}\big(f_{\mathring{\phi}}( \boldsymbol{G} )\big), g_{\mathring{\varphi}}\big(f_{\mathring{\phi}}(\widehat{\boldsymbol{G}^\prime} )\big) \Big) \bigg)
	\end{aligned}
	\label{eq:fastweightsigma}
\end{equation}
where $\ell^\prime$ represents the learning rate of $\sigma$, and $\widehat{\boldsymbol{G}^\prime}$ is the augmented graphs with stop-gradient, which is defined as $\widehat{\boldsymbol{G}^\prime} = a_\sigma(\boldsymbol{G}).detach()$. $\mathcal{L}_{MEGA}$ is to train $a_{\sigma}(\cdot)$ to generate hard and informative augmented graphs defined as follows:
\begin{equation}
    \begin{aligned}
    	 \mathcal{L}_{MEGA} = \underbrace{tr(\boldsymbol{C}) - de(\boldsymbol{C})}_{instance \ term} + \lambda \underbrace{\left(tr(|\mathbb{1} - \boldsymbol{D}|^2) + de( | \boldsymbol{D}|^2)\right)}_{feature \ term}
    \label{eq:mega}
    \end{aligned}
\end{equation}
where $tr(\cdot)$ denotes the matrix trace function, which is defined as $tr(M) = \sum_{i}M_{ii}$, and $de(\cdot)$ is a matrix calculation function defined as $de(M) = \sum_i\sum_{j \neq i}M_{ij}$. $|\cdot|^2$ presents a matrix element-wise square function defined as $|M|^2 = M \times M$ by Hadamard product, and $\mathbb{1}$ presents an identity matrix. $\lambda$ is the coefficient that controls the balance between two terms of $\mathcal{L}_{MEGA}$. Intuitively, the $instance \ term$ aims to lead MEGA to generate instance-level challenging examples for self-supervised learning. Inspired by \cite{2021Barlow}, we design the $feature \ term$ to promote the model to learn dimensionally non-redundant representations, respectively. Concretely, minimizing the proposed $\mathcal{L}_{MEGA}$ by using the second-derivative technique can guide $a_{\sigma}(\cdot)$ to generate \textit{hard} and \textit{informative} augmented graphs. $\boldsymbol{C}$ denotes the cross-correlation matrix computed between the features of the original graphs and augmented graphs in a batch, as follows:
\begin{equation}
    \begin{aligned}
    	 \boldsymbol{C}_{ij} = \frac{\boldsymbol{z}_i \cdot \boldsymbol{z}_j^\prime}{|\boldsymbol{z}_i| \cdot |\boldsymbol{z}^\prime_j|}
    \label{eq:corc}
    \end{aligned}
\end{equation}
where $i, j \in \llbracket{1, N} \rrbracket$ in a batch $N$ of graphs. $\boldsymbol{D}$ is the cross-correlation matrix computed between the multi-dimensional features of the original graphs and the corresponding augmented graphs along the batch, which is defined as:
\begin{equation}
    \begin{aligned}
    	 \boldsymbol{D}_{pq} = \frac{\sum_i(\boldsymbol{z}_{i,p} \cdot \boldsymbol{z}_{i,q}^\prime)}{\sqrt{\sum_i(\boldsymbol{z}_{i,p})^2} \cdot \sqrt{\sum_i(\boldsymbol{z}^\prime_{i,q})^2}}
    \label{eq:corc}
    \end{aligned}
\end{equation}
where $i \in \llbracket{1, N} \rrbracket$ indexes batch graphs and $p, q \in \llbracket{1, N^D} \rrbracket$ index the feature dimension of the original graph and the corresponding augmented graph, and $N^D$ denotes the number of feature dimension. $\boldsymbol{C}$, $\boldsymbol{D}$ aim to train $a_{\sigma}(\cdot)$ to generate \textit{hard} and \textit{informative} augmented graphs, respectively. Concretely, the objective of auxiliary meta-learning is to enable LGA to learn augmented graphs that are hard and informative for the \textit{encoder}, thereby improving the encoder's learning process for the \textit{next} iteration.

\section{Experiments}

\begin{figure*}[ht]
	\centering
	\subfigure[MUTAG]{
		\begin{minipage}{0.3\textwidth} 
			\includegraphics[width=\textwidth]{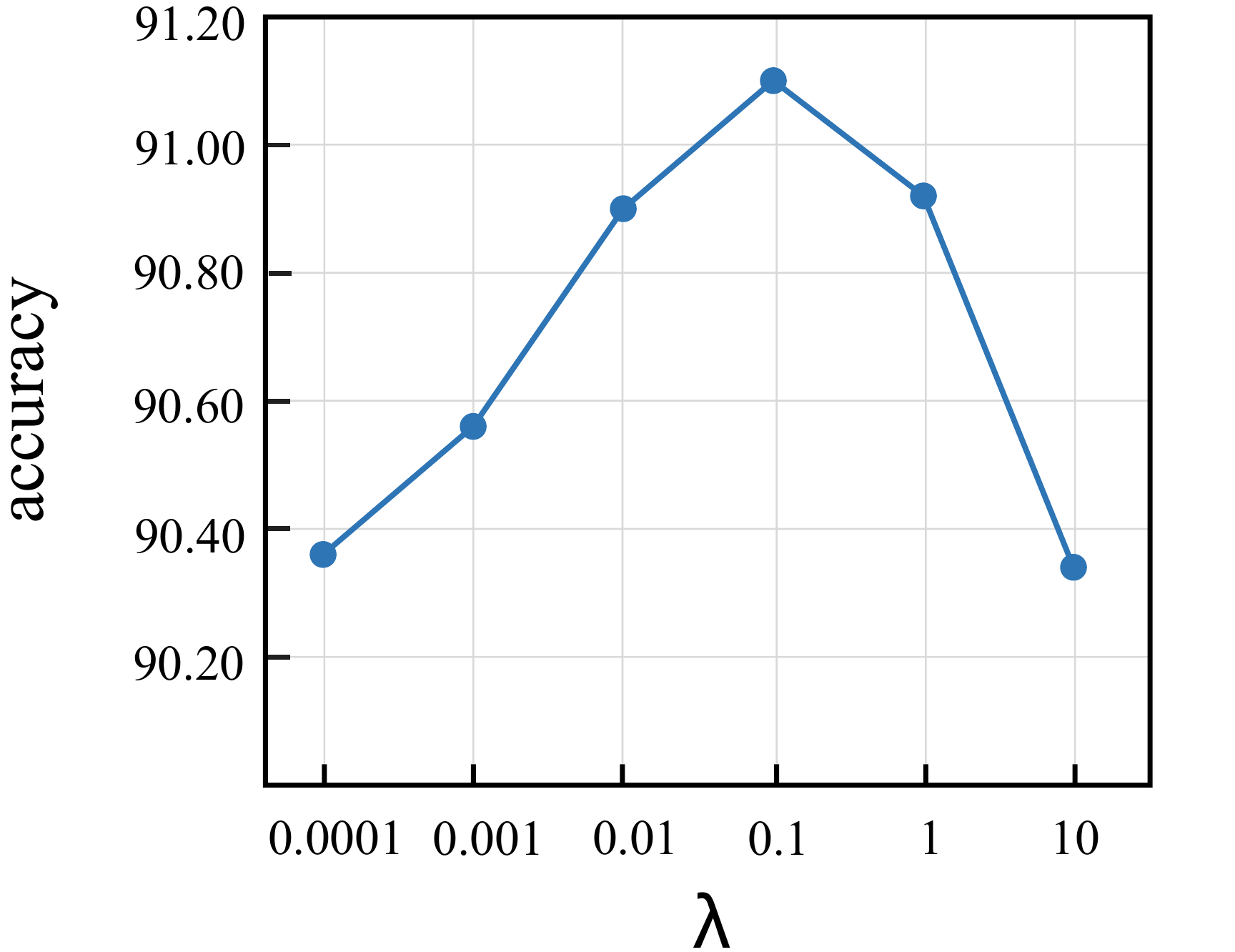} \\
		\end{minipage}
	}
	\subfigure[PROTEINS]{
		\begin{minipage}{0.3\textwidth}
			\includegraphics[width=\textwidth]{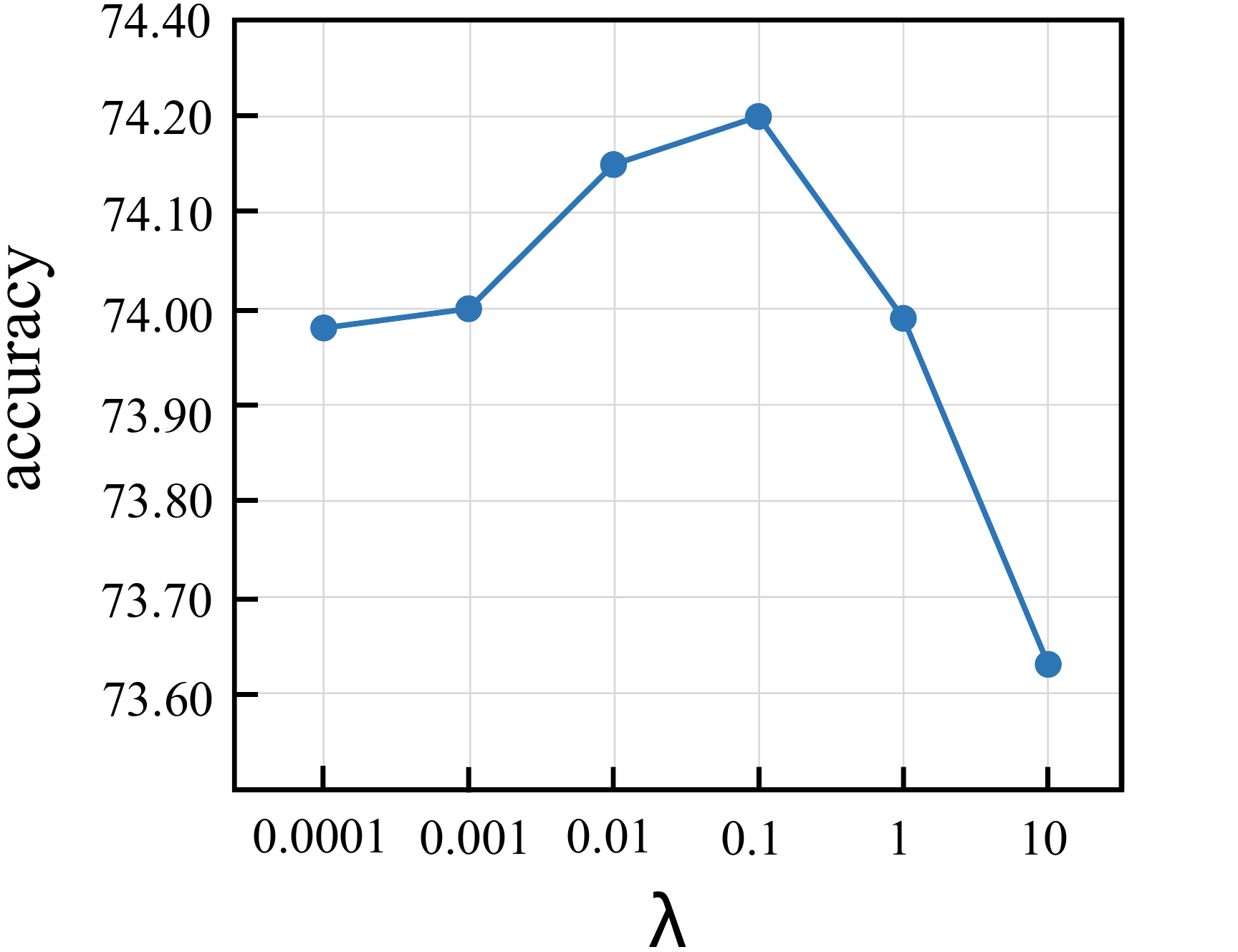} \\
		\end{minipage}
	}
	\subfigure[IMDB-B]{
		\begin{minipage}{0.3\textwidth}
			\includegraphics[width=\textwidth]{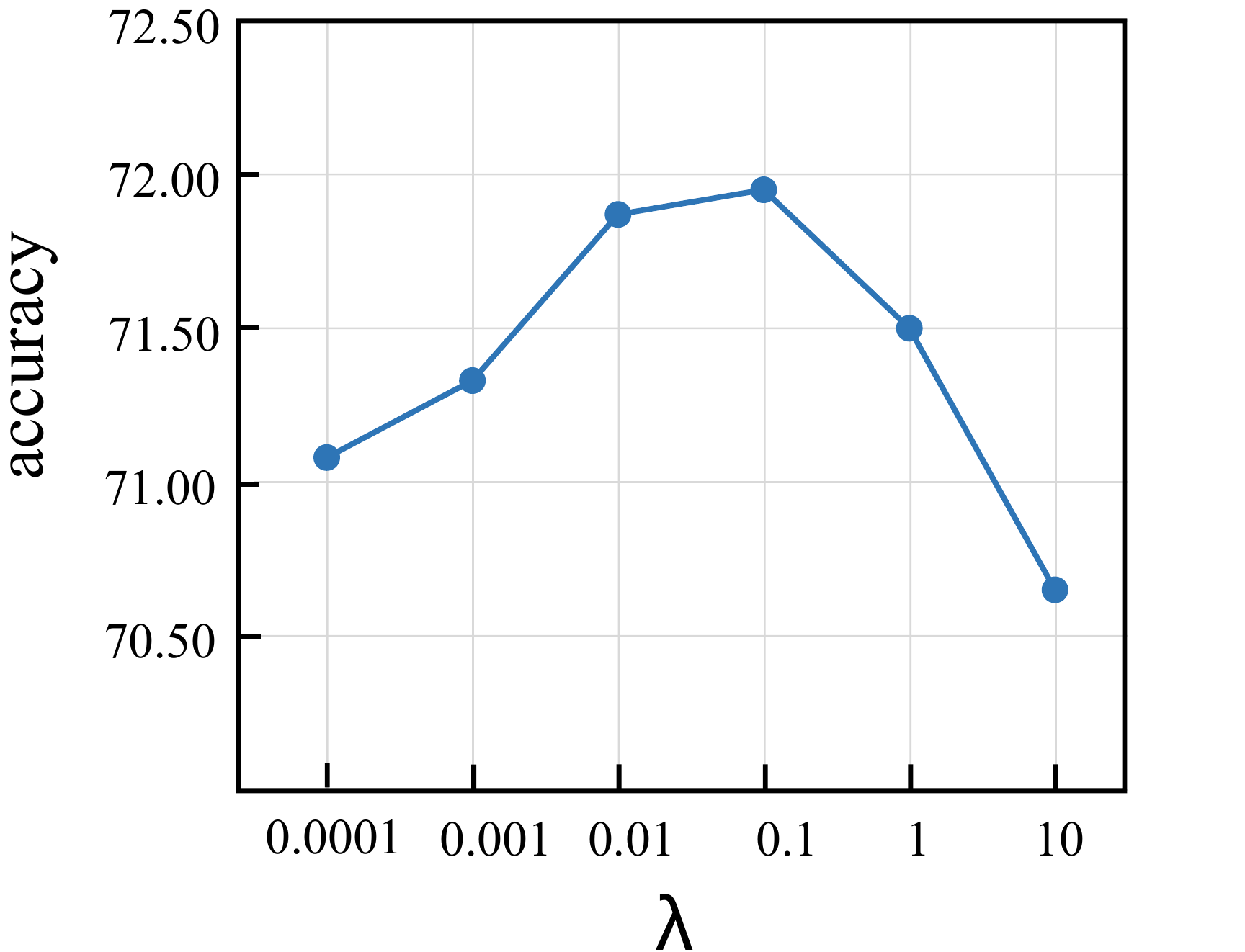} \\
		\end{minipage}
	}
	\caption{Results of MEGA's performance with a range of factor $\lambda$. We perform MEGA on three benchmark datasets: MUTAG, PROTEINS, and IMDB-B. The abscissa axis represents the value of $\lambda$, and the ordinate axis represents the accuracies.}
	\label{fig:blt}
\end{figure*}

\begin{figure*}[ht]
	\centering
	\includegraphics[width=1.0\textwidth]{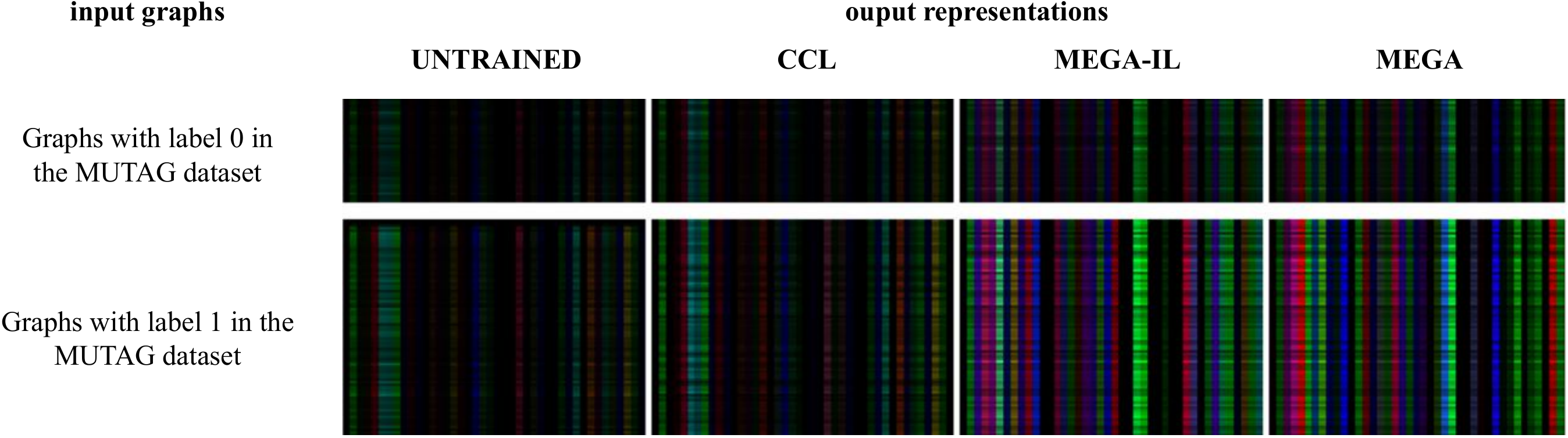}
	\caption{This figure shows the visualized output graph features on the MUTAG dataset. The graph features are projected into a color image in RGB format, where different colors represent different types of features. The abscissa axis represents the output feature dimensions of compared methods, and the ordinate axis represents graphs of different classes.}
	\label{fig:visual}
\end{figure*}

In this section, we demonstrate the effectiveness of MEGA on various benchmark datasets. Our experiments were conducted in an unsupervised learning setting.

\subsection{Comparison with State-of-the-art Methods}
\textbf{Datasets.} We evaluate our method on twelve benchmark datasets in two major categories: 1) Social Networks: RDT-M5K, IMDB-B, IMDB-M from TU Dataset \cite{morris2020tudataset}. 2) Molecules: PROTEINS, MUTAG, COLLAB and DD from TU Dataset \cite{morris2020tudataset} and molesol, mollipo, molbbbp, moltox21 and molsider from Open Graph Benchmark (OGB) \cite{hu2020open}. 

\paragraph{Experiment settings.} We compared MEGA with four unsupervised/self-supervised learning baselines, which include randomly initialized untrained GIN (GIN RIU) \cite{xu2018powerful}, InfoGraph \cite{sun2020infograph}, GraphCL \cite{you2020graph} and AD-GCL \cite{suresh2021adversarial}. Experiment results of InfoGraph \cite{sun2020infograph} and GraphCL \cite{you2020graph} show that they generally outperform graph kernel and network embedding methods including \cite{kriege2020survey}, \cite{grover2016node2vec}, and \cite{adhikari2018sub2vec}
. As discussed in the method section, the instance-level constrains and feature-level constraints of $\mathcal{L}_{MEGA}$ are balanced by parameter $\lambda$. To study the effects of these constraints, we set $\lambda=0$ for the ablation study, termed MEGA-IL. We followed the experimental protocol of AD-GCL, including the train/validation/test splits. The average classification accuracy with standard deviation on the test results over the last ten runs of training is reported. For a fair comparison, we adopted GIN as the encoder as other baselines do. We adopt the Adam optimizer with a learning rate of $10^{-4}$ for learnable graph augmentation and a learning rate of $10^{-3}$ for graph encoding. We use 50 training epochs on all datasets. All methods adopt a downstream linear classifier or regressor with the same hyper-parameters.

\paragraph{Results.} The results are reported in Table \ref{tab:test_tu} and \ref{tab:test_ogbg}. The results show that MEGA achieves the best results compared with baselines across benchmark datasets. We attribute such performance to MEGA's abilities to generate both hard and informative augmented graph features. The results show that MEGA outperforms MEGA-IL across most datasets, which proves that the feature-level constraints do improve the network to learn informative representations. MEGA-IL still performs better than most of the baselines that adopt the same encoder and contrastive learning pattern, which means that the instance-level constrains of $\mathcal{L}_{MEGA}$ work well.

\subsection{Evaluation of Feature-level Constrains}
For further evaluation of feature-level constraints, we change the value of $\lambda$ and observe how the performance changes. We adopt three different datasets, including two molecule datasets and one social network dataset. 

The results are reported in Figure \ref{fig:blt}. The performance changes as the factor $\lambda$ changes. When $\lambda$ takes 0.1, the performance is optimal among all tasks. The results prove that feature-level constraints can enhance the discrimination of features to a certain extent. The feature-level constraints ensure that the generated augmented graph correlates with the original graph, preventing LGA from learning outrageous graph augmentation. However, if we overly increase the impact of feature-level constraints, the generation of hard augmented graph features could be interfered.

\subsection{Analyze on Representations}
To better understand the quality of the representations learned by MEGA, we visualize the output graph features. For comparison, we conducted experiments on four different networks: (1) UNTRAINED, a randomly initialized GIN encoder without training. (2) CCL, a conventional graph contrastive learning network without our proposed LGA. (3) MEGA-IL, MEGA without feature-level constraints. (4) MEGA. 

The results are shown in Figure \ref{fig:visual}. We intuitively observe the representations of each graph and find that MEGA and MEGA-IL output more "colorful" results than RI-GIN and CCL, which indicates that their output is more informative. In detail, for MEGA and MEGA-IL, there are many vertical lines of different colors, which means that the difference between the dimensions of the feature is significant. This phenomenon is more evident on MEGA, indicating that the feature-level constraints make the feature dimensions less redundant. 


\section{Conclusions}
This paper proposed a novel meta graph augmentation to boost the representation ability of graph contrastive learning. We apply secondary derivative technique to update a learnable graph augmenter, which is to generate \textit{hard} and \textit{informative} augmented graph for contrastive learning. This way, we can yield a representation with uniformity at the instance-level and informativeness at the feature-level.


\section*{Acknowledgements}
The authors thank all the anonymous reviewers. This work is supported by the Strategic Priority Research Program of the Chinese Academy of Sciences, Grant No. XDA19020500.

\bibliographystyle{named}
\bibliography{mybibfile}

\begin{thebibliography}{}

\bibitem[\protect\citeauthoryear{Adhikari \bgroup \em et al.\egroup
  }{2018}]{adhikari2018sub2vec}
Bijaya Adhikari, Yao Zhang, Naren Ramakrishnan, and B~Aditya Prakash.
\newblock Sub2vec: Feature learning for subgraphs.
\newblock In {\em Pacific-Asia Conference on Knowledge Discovery and Data
  Mining}, 2018.

\bibitem[\protect\citeauthoryear{Chen \bgroup \em et al.\egroup
  }{2020}]{chen2020simple}
Ting Chen, Simon Kornblith, Mohammad Norouzi, and Geoffrey Hinton.
\newblock A simple framework for contrastive learning of visual
  representations.
\newblock In {\em International conference on machine learning}. PMLR, 2020.

\bibitem[\protect\citeauthoryear{Chuang \bgroup \em et al.\egroup
  }{2020}]{chuang2020debiased}
Ching-Yao Chuang, Joshua Robinson, Lin Yen-Chen, Antonio Torralba, and Stefanie
  Jegelka.
\newblock Debiased contrastive learning.
\newblock {\em arXiv preprint arXiv:2007.00224}, 2020.

\bibitem[\protect\citeauthoryear{Ermolov \bgroup \em et al.\egroup
  }{2021}]{2020WhiteningErmolov}
Aleksandr Ermolov, Aliaksandr Siarohin, Enver Sangineto, and Nicu Sebe.
\newblock Whitening for self-supervised representation learning.
\newblock In {\em International Conference on Machine Learning}, pages
  3015--3024. PMLR, 2021.

\bibitem[\protect\citeauthoryear{Finn \bgroup \em et al.\egroup
  }{2017}]{2017ModelFinn}
Chelsea Finn, Pieter Abbeel, and Sergey Levine.
\newblock Model-agnostic meta-learning for fast adaptation of deep networks.
\newblock In {\em Proceedings of the 34th ICML}. {PMLR}, 2017.

\bibitem[\protect\citeauthoryear{Grill \bgroup \em et al.\egroup
  }{2020}]{2020Bootstrap}
Jean-Bastien Grill, Florian Strub, Florent Altch{\'e}, Corentin Tallec, Pierre
  Richemond, Elena Buchatskaya, Carl Doersch, Bernardo Avila~Pires, Zhaohan
  Guo, Mohammad Gheshlaghi~Azar, et~al.
\newblock Bootstrap your own latent-a new approach to self-supervised learning.
\newblock {\em Advances in Neural Information Processing Systems},
  33:21271--21284, 2020.

\bibitem[\protect\citeauthoryear{Grover and
  Leskovec}{2016}]{grover2016node2vec}
Aditya Grover and Jure Leskovec.
\newblock node2vec: Scalable feature learning for networks.
\newblock In {\em Proceedings of the 22nd ACM SIGKDD international conference
  on Knowledge discovery and data mining}, pages 855--864, 2016.

\bibitem[\protect\citeauthoryear{Hassani and
  Khasahmadi}{2020}]{hassani2020contrastive}
Kaveh Hassani and Amir~Hosein Khasahmadi.
\newblock Contrastive multi-view representation learning on graphs.
\newblock In {\em International Conference on Machine Learning}, pages
  4116--4126. PMLR, 2020.

\bibitem[\protect\citeauthoryear{Hu \bgroup \em et al.\egroup
  }{2020a}]{hu2020strategies}
W~Hu, B~Liu, J~Gomes, M~Zitnik, P~Liang, V~Pande, and J~Leskovec.
\newblock Strategies for pre-training graph neural networks.
\newblock In {\em International Conference on Learning Representations (ICLR)},
  2020.

\bibitem[\protect\citeauthoryear{Hu \bgroup \em et al.\egroup
  }{2020b}]{hu2020open}
Weihua Hu, Matthias Fey, Marinka Zitnik, Yuxiao Dong, Hongyu Ren, Bowen Liu,
  Michele Catasta, and Jure Leskovec.
\newblock Open graph benchmark: Datasets for machine learning on graphs.
\newblock {\em Neural Information Processing Systems (NeurIPS)}, 2020.

\bibitem[\protect\citeauthoryear{Kipf and Welling}{2016}]{kipf2016semi}
Thomas~N Kipf and Max Welling.
\newblock Semi-supervised classification with graph convolutional networks.
\newblock {\em arXiv preprint arXiv:1609.02907}, 2016.

\bibitem[\protect\citeauthoryear{Kriege \bgroup \em et al.\egroup
  }{2020}]{kriege2020survey}
Nils~M Kriege, Fredrik~D Johansson, and Christopher Morris.
\newblock A survey on graph kernels.
\newblock {\em Applied Network Science}, 5(1):1--42, 2020.

\bibitem[\protect\citeauthoryear{Liu \bgroup \em et al.\egroup
  }{2019}]{2019SelfLiu}
Shikun Liu, Andrew Davison, and Edward Johns.
\newblock Self-supervised generalisation with meta auxiliary learning.
\newblock {\em Advances in Neural Information Processing Systems}, 32, 2019.

\bibitem[\protect\citeauthoryear{Morris \bgroup \em et al.\egroup
  }{}]{morris2020tudataset}
Christopher Morris, Nils~M Kriege, Franka Bause, Kristian Kersting, Petra
  Mutzel, and Marion Neumann.
\newblock Tudataset: A collection of benchmark datasets for learning with
  graphs.
\newblock In {\em ICML 2020 Workshop on Graph Representation Learning and
  Beyond}.

\bibitem[\protect\citeauthoryear{Oord \bgroup \em et al.\egroup
  }{2018}]{2018RepresentationOord}
Aaron van~den Oord, Yazhe Li, and Oriol Vinyals.
\newblock Representation learning with contrastive predictive coding.
\newblock {\em arXiv preprint arXiv:1807.03748}, 2018.

\bibitem[\protect\citeauthoryear{Schmidhuber}{2014}]{2014Schmidhuber}
J{\"u}rgen Schmidhuber.
\newblock Learning complex, extended sequences using the principle of history
  compression.
\newblock {\em Neural Computation}, 4(2):234--242, 2014.

\bibitem[\protect\citeauthoryear{Sun \bgroup \em et al.\egroup
  }{2020}]{sun2020infograph}
Fan-Yun Sun, Jordon Hoffman, Vikas Verma, and Jian Tang.
\newblock Infograph: Unsupervised and semi-supervised graph-level
  representation learning via mutual information maximization.
\newblock In {\em International Conference on Learning Representations}, 2020.

\bibitem[\protect\citeauthoryear{Suresh \bgroup \em et al.\egroup
  }{2021}]{suresh2021adversarial}
Susheel Suresh, Pan Li, Cong Hao, and Jennifer Neville.
\newblock Adversarial graph augmentation to improve graph contrastive learning.
\newblock {\em Advances in Neural Information Processing Systems}, 34, 2021.

\bibitem[\protect\citeauthoryear{Tian \bgroup \em et al.\egroup
  }{2020}]{tian2020contrastive}
Yonglong Tian, Dilip Krishnan, and Phillip Isola.
\newblock Contrastive multiview coding.
\newblock In {\em Computer Vision--ECCV 2020: 16th European Conference,
  Glasgow, UK, August 23--28, 2020, Proceedings, Part XI 16}, pages 776--794.
  Springer, 2020.

\bibitem[\protect\citeauthoryear{Veli{\v{c}}kovi{\'c} \bgroup \em et al.\egroup
  }{2017}]{velivckovic2017graph}
Petar Veli{\v{c}}kovi{\'c}, Guillem Cucurull, Arantxa Casanova, Adriana Romero,
  Pietro Lio, and Yoshua Bengio.
\newblock Graph attention networks.
\newblock {\em arXiv preprint arXiv:1710.10903}, 2017.

\bibitem[\protect\citeauthoryear{Veli{\v{c}}kovi{\'c} \bgroup \em et al.\egroup
  }{2018}]{velivckovic2018deep}
Petar Veli{\v{c}}kovi{\'c}, William Fedus, William~L Hamilton, Pietro Li{\`o},
  Yoshua Bengio, and R~Devon Hjelm.
\newblock Deep graph infomax.
\newblock {\em arXiv preprint arXiv:1809.10341}, 2018.

\bibitem[\protect\citeauthoryear{Wan \bgroup \em et al.\egroup
  }{2020}]{wan2020contrastive}
Sheng Wan, Shirui Pan, Jian Yang, and Chen Gong.
\newblock Contrastive and generative graph convolutional networks for
  graph-based semi-supervised learning.
\newblock {\em arXiv preprint arXiv:2009.07111}, 2020.

\bibitem[\protect\citeauthoryear{Xu \bgroup \em et al.\egroup
  }{2018}]{xu2018powerful}
Keyulu Xu, Weihua Hu, Jure Leskovec, and Stefanie Jegelka.
\newblock How powerful are graph neural networks?
\newblock In {\em International Conference on Learning Representations}, 2018.

\bibitem[\protect\citeauthoryear{You \bgroup \em et al.\egroup
  }{2020}]{you2020graph}
Yuning You, Tianlong Chen, Yongduo Sui, Ting Chen, Zhangyang Wang, and Yang
  Shen.
\newblock Graph contrastive learning with augmentations.
\newblock {\em Advances in Neural Information Processing Systems},
  33:5812--5823, 2020.

\bibitem[\protect\citeauthoryear{Zbontar \bgroup \em et al.\egroup
  }{2021}]{2021Barlow}
Jure Zbontar, Li~Jing, Ishan Misra, Yann LeCun, and St{\'e}phane Deny.
\newblock Barlow twins: Self-supervised learning via redundancy reduction.
\newblock In {\em International Conference on Machine Learning}, pages
  12310--12320. PMLR, 2021.

\end{thebibliography}

\end{document}